\documentclass{article}

\PassOptionsToPackage{numbers, compress}{natbib}

\usepackage[final]{neurips_2021}

\PassOptionsToPackage{hyphens}{url}

\usepackage[utf8]{inputenc} 
\usepackage[T1]{fontenc}    
\usepackage{url}            
\usepackage{hyperref}       
\usepackage{booktabs}       
\usepackage{amsfonts}       
\usepackage{nicefrac}       
\usepackage{microtype}      
\usepackage{xcolor}         
\usepackage{multirow}
\usepackage{multicol}
\usepackage{amsmath}
\usepackage{algorithm}
\usepackage[noend]{algpseudocode}
\usepackage{wrapfig}
\usepackage{graphicx}
\usepackage{caption}
\usepackage{pdflscape}

\newcommand\norm[1]{\left\lVert#1\right\rVert}
\DeclareUnicodeCharacter{2212}{-}
\renewcommand{\arraystretch}{1.1}
\usepackage{titlesec}

\titlespacing\section{0pt}{8pt plus 4pt minus 2pt}{0pt plus 2pt minus 2pt}
\titlespacing\subsection{0pt}{12pt plus 4pt minus 2pt}{0pt plus 2pt minus 2pt}
\titlespacing\subsubsection{0pt}{12pt plus 4pt minus 2pt}{0pt plus 2pt minus 2pt}

\title{Generative Optimization Networks for Memory Efficient Data Generation}

\author{%
  Shreshth Tuli\\
  Imperial College London\\
  \texttt{s.tuli20@imperial.ac.uk} \\
  \And%
  Shikhar Tuli\\
  Princeton University\\
  \texttt{stuli@princeton.edu} \\
   \AND%
   Giuliano Casale\\
   Imperial College London \\
   \texttt{g.casale@imperial.ac.uk} \\
   \And%
   Nicholas R. Jennings\\
   Loughborough University \\
   \texttt{n.r.jennings@lboro.ac.uk} \\
}

\begin{document}

\maketitle

\begin{abstract}
In standard generative deep learning models, such as autoencoders or GANs, the size of the parameter set is proportional to the complexity of the generated data distribution. A significant challenge is to deploy resource-hungry deep learning models in devices with limited memory to prevent system upgrade costs. To combat this, we propose a novel framework called generative optimization networks (GON) that is similar to GANs, but does not use a generator, significantly reducing its memory footprint. GONs use a single discriminator network and run optimization in the input space to generate new data samples, achieving an effective compromise between training time and memory consumption. GONs are most suited for data generation problems in limited memory settings. Here we illustrate their use for the problem of anomaly detection in memory-constrained edge devices arising from attacks or intrusion events. Specifically, we use a GON to calculate a reconstruction-based anomaly score for input time-series windows. Experiments on a Raspberry-Pi testbed with two existing and a new suite of datasets show that our framework gives up to 32\% higher detection F1 scores and 58\% lower memory consumption, with only 5\% higher training overheads compared to the state-of-the-art.
\end{abstract}

\section{Introduction}
\label{sec:intro}
The past years have witnessed widespread adoption of deep generative models in cyber-physical systems~\cite{sharma2019anomaly}. This is primarily due to the rise of affordable and powerful computation in the Internet of Things (IoT) era~\cite{kim2017review}. Recently, such neural models have become increasingly demanding in terms of their memory requirements, resulting in many challenging problems. A canonical example, that we will use throughout this paper, is anomaly detection in  resource-constrained edge systems that place devices close to the end user for low latency service delivery~\cite{mad_gan}. Such systems generate enormous amounts of multivariate time-series data in the form of system device information or the user inputs~\cite{kim2017review, mad_gan}. As applications become more demanding and privacy sensitive, edge devices are more prone to breakdowns, malicious attacks and intrusions~\cite{zhang2019serious}. Thus, it has become crucial to maintain resilience in edge systems. As sending all system operation data to a cloud backend is infeasible, the challenge is to efficiently run accurate system anomaly detection at the edge~\cite{cae_m}.


\textbf{Challenges.} The problem of deploying generative models in memory-constrained devices is hard. For example, detection of anomalies at the edge where devices typically have at most 8GB of RAM available~\cite{mad_gan}. Moreover, battery-operated hardware makes it harder to run heavy analysis models such as deep autoencoders~\cite{khanna2020intelligent}. In such systems, it is possible to handle the processing limitations by effective preemption and prolonged job execution. However, memory bottlenecks are much harder to solve due to the limited performance of virtual memory in network attached storage settings~\cite{shao2020communication}. 

\textbf{Existing Solutions.} In recent years, many anomaly detection methods have been proposed that utilize deep autoencoders (more details in Section~\ref{sec:related}). However, the models that have the best detection scores often have a very high memory footprint (4-6 GB for training and inference in state-of-the-art approaches). Some prior work aims at reducing the on-device parameter size of such models using techniques like model compression~\cite{mc}, neural network slimming~\cite{gan_slimming} or model distribution~\cite{li2021model}. However, such approaches usually have performance penalties associated with them.

\textbf{Our Contributions.} We propose a new framework called generative optimization networks (GON).\footnote{The code and datasets are available on GitHub at \url{https://github.com/imperial-qore/GON}.} This is similar to generative adversarial nets (GAN)~\cite{goodfellow2014generative}, but without the generator. The novel insight of our work is that a sufficiently trained discriminator should not only indicate whether an input belongs to a data distribution, but also how to tweak the input to make it resemble the target distribution more closely. This can be done using the gradient of the discriminator output with respect to its input~\cite{tuli2021cosco} and executing gradient optimization of the output score. Having only a single network should give significant memory footprint gains, at the cost of higher training times (as generation would entail executing optimization in the input space). This is acceptable in many applications scenarios where initial model training is only a one-time activity~\cite{goli2020migrating, huang2020clio}. The GON framework is agnostic to the specific use case and can be utilized in any application where GANs are used. In this article, we explore the important use case of anomaly detection in memory-constrained edge devices, specifically Raspberry Pis. A broader impact statement is given in Appendix~\ref{sec:broader_impact}.

\section{Related Work}
\label{sec:related}
In the past, several unsupervised methods have been proposed that are able to detect anomalies in multivariate time-series data~\cite{mad_gan, topomad, if, dilof, ocsvm, usad}. Among these, the traditional models include ensemble based Isolation Forest (IF)~\cite{if}, memory efficient local outlier detection (DILOF)~\cite{dilof} and an extended Kalman filter based one-class SVM anomaly detector (OCSVM)~\cite{ocsvm}. Such models are known to be quick in anomaly detection, but can incur low accuracy for multi-dimensional and complex data~\cite{usad}. We compare GON against IF, DILOF and OCSVM as baselines.

Most state-of-the-art anomaly detection systems use deep learning based solutions for pattern recognition and discern deviations for such patterns. Typically, such techniques divide the time-series data into discrete time-series windows and create a reconstruction of such window inputs. The deviation of the reconstructed window from the input is used to generate an anomaly score, a continuous running value that indicates the likelihood of an anomaly. The anomaly prediction labels are generated using one of the various thresholding techniques.  A category within these techniques uses light-weight online sequential extreme learning machines (OS-ELM) for their fast learning. One such memory-efficient technique of edge devices is the on-device learning anomaly detector, ONLAD~\cite{onlad} that uses a combination of OS-ELM and autoencoders to generate window reconstructions. Other methods use generative models like variational autoencoders or GANs for the same task and have been shown to perform very well~\cite{usad, mad_gan}. However, to fit such models in memory-constrained environments, we use model compression and GAN slimming respectively~\cite{mc, gan_slimming}, with their respective precision tradeoffs. Thus, our baselines include MAD-GAN with GAN Slimming~\cite{mad_gan, gan_slimming}, autoencoder based USAD with model compression~\cite{usad,mc} and convolutional networks with CNN Slimming, CAE-M~\cite{cae_m, cnn_slimming}. We also include other models with flexible model sizes like fully connected network based SlimGAN~\cite{slimgan}. 

\section{Generative Optimization Networks}

\begin{figure}[t]
    \centering
    \includegraphics[width=\linewidth]{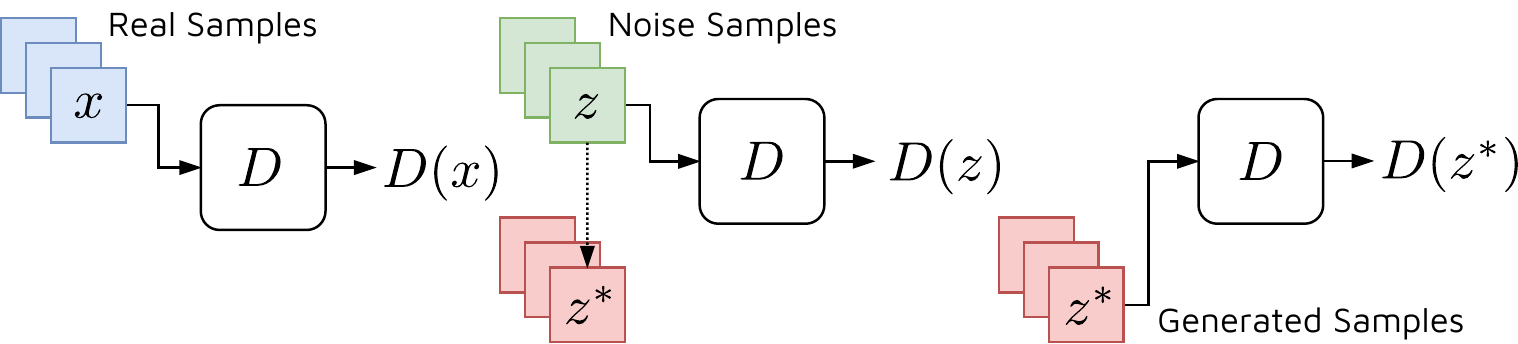}
    \caption{The GON framework.}
    \label{fig:model}
\end{figure}

In the Generative Optimization Networks framework, we have a single discriminator model $D(\cdot ; \theta)$ that is a differentiable multilayer perceptron with parameters $\theta$. For any input $x$, the output $D(x; \theta)$ is a scalar. We denote a generated distribution with $p_g$, and with $D(x; \theta)$ the probability that $x$ belongs to the data $p_{data}$ instead of $p_g$. We train $D$ such that it correctly labels samples from data and $p_g$. 
A working representation is shown in Fig.~\ref{fig:model}. The key idea is divided into three parts.

\textbf{Training using real samples.} For any sample $x$ from the data, to make sure the output of $D$ is maximized, we train it to minimize the cross-entropy loss by descending the stochastic gradient
\begin{equation}
\label{eq:l1}
    - \nabla_\theta \log \big(D(x; \theta) \big).
\end{equation}

\textbf{Generating new samples.} Starting from a random noise sample $z$, we aim to increase its probability of coming from the data. To do this, we leverage neural network inversion~\cite{jensen1999inversion}, specifically using $D$ as a surrogate model and maximize the objective $\log \big(D(z) \big)$. To do this, starting from a random noise sample, a stochastic gradient ascent based solution executes the following until convergence
\begin{equation}
\label{eq:opt}
    z \gets z + \gamma \nabla_z \log \big( D(z; \theta) \big).
\end{equation}
Here, $\gamma$ is a step parameter. 
The updates can be improved to reduce convergence times by using optimizers like \texttt{Adam} or its variants with warm restarts using cosine annealing~\cite{loshchilov2018decoupled}. The optimization loop converges to $z^*$ that should have a high $D(z; \theta)$ probability. The working intuition is that the surrogate surface is highly non-convex and optimization over diverse noise samples would lead to distinct local optima. In principle, every optimum corresponds to a unique element from the data distribution, allowing the GON model to generate a diverse set of samples. Alternatively, we could stop after each step with  probability $D(z)$ (see Appendix~\ref{sec:broader_impact}). A caveat of this approach is that sample generation could be slow as we run optimization in the input space. For some applications the overhead might be negligible or acceptable~\cite{goli2020migrating, huang2020clio}.

\textbf{Training using generated samples.} We consider the new samples as self-supervised fake ones~\cite{chen2020self} not belonging to the data and train $D$ using the cross-entropy loss by descending the gradient
\begin{equation}
\label{eq:l2}
     - \nabla_\theta \log \big(1 - D(z^*; \theta) \big).
\end{equation}

Informally, as $D$ becomes a better discriminator, minimizing~\eqref{eq:l1} and~\eqref{eq:l2}, the generated samples should also be close to the data. 
A more pragmatic implementation of the GON framework is shown in Alg.~\ref{alg:gon}, where we combine the first and last steps when training $D$ using minibatches.

\begin{algorithm}[!t]
    \begin{algorithmic}[1]
    \Require
    \For{ number of training iterations }
    \State Sample minibatch of size $m$ noise samples $\{z^{(1)}, \ldots, z^{(m)}\}$.
    \State Sample minibatch of size $m$ samples $\{x^{(1)}, \ldots, x^{(m)}\}$ from given data.
    \State Generate new samples $\{z^{*(1)}, \ldots, z^{*(m)}\}$ by running the following till convergence
    \begin{equation*}
        z \gets z + \gamma \nabla_z \log \big(D(z; \theta) \big).
    \end{equation*} \label{line:optimize}
    \State Update the discriminator $D(\cdot; \theta)$ by ascending the stochastic gradient.
    \begin{equation*}
        \nabla_\theta \frac{1}{m} \sum_{i=1}^m \big[\log \big(D(x^{(i)}; \theta) \big) + \log \big(1 - D(z^{*(i)}) \big) \big].
    \end{equation*} \label{line:train}
    \EndFor
    \end{algorithmic}
\caption{Minibatch stochastic gradient based training of generative optimization networks. Input is dataset $\Lambda$ and hyperparameters $m$ and $\gamma$. In experiments we use \texttt{Adam} optimizer for lines~\ref{line:optimize} and~\ref{line:train}.}
\label{alg:gon}
\end{algorithm}

\section{Anomaly Detection using a GON}
For unsupervised anomaly detection using the GON framework, for each input datum from a multivariate time-series, we create a reconstruction using a neural network $D$. To capture the temporal context, we consider that the time-series data is converted into local contextual sliding windows of length $K$. Thus, a datapoint at time $t$ is represented as $x_t$ as a sequence of $K$ elements. Now, if $t$ varies from $0$ to $T$, we have a dataset $\Lambda = \{x_0, \ldots, x_t\}$ created using replication padding for $t < K$. To train $D$, the real samples are obtained from $\Lambda$ and fake samples using the \texttt{Adam} on~\eqref{eq:opt} starting from a randomly initialized window and the model is trained using Alg.~\ref{alg:gon}.

Now, for an input window $x_t$, we create a reconstruction $\hat{x}_t$, using a trained model $D$ and the \texttt{Adam} based version of~\eqref{eq:opt}. The reconstruction error between the last elements is used as an anomaly score and denoted by 
\begin{equation}
    S(x_t) = \norm{x_t - \hat{x}_t}.
\end{equation}

Once we have the anomaly scores, we generate anomaly labels using the Peak Over Threshold (POT) method~\cite{siffer2017anomaly}. POT chooses the threshold automatically and dynamically, above which an anomaly score is classified as a true class label.

\section{Experiments}
\label{sec:experiments}

\textbf{Baselines.} We compare the GON model against baselines: OCSVM~\cite{ocsvm}, IF~\cite{if}, DILOF~\cite{dilof}, ONLAD~\cite{onlad}, CAE-M with CNN slimming~\cite{cae_m}, USAD with model compression~\cite{usad}, MAD-GAN with GAN Slimming~\cite{mad_gan} and SlimGAN~\cite{slimgan} (more details in Section~\ref{sec:related}). 


\begin{table}
	\begin{minipage}{0.60\linewidth}
	    \centering
        \caption{Dataset Statistics}
        \begin{tabular}{@{}llrrrr@{}}
        \toprule
        Dataset &  & Train & Test & Dimensions & Anomalies (\%)\tabularnewline
        \midrule
        FTSAD-1 &  & 600 & 5000 & 1 & 12.88\tabularnewline
        FTSAD-25 &  & 574 & 1700 & 25 & 32.23\tabularnewline
        FTSAD-55 &  & 2158 & 2264 & 55 & 13.69\tabularnewline
        SMD &  & 5696 & 5696 & 38 & 11.32\tabularnewline
        MSDS &  & 4740 & 4740 & 10 & 5.93\tabularnewline
        \bottomrule
        \end{tabular}
        \label{tab:datasets}
	\end{minipage}\hfill
	\begin{minipage}{0.35\linewidth}
        \centering \setlength{\belowcaptionskip}{-9pt}
        \includegraphics[width=\linewidth]{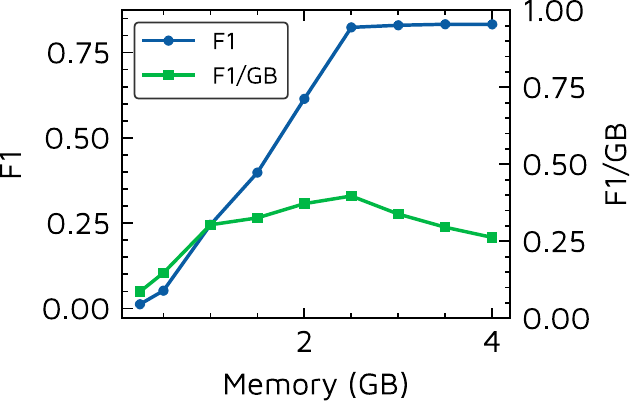}
        \captionof{figure}{\footnotesize F1 and F1/GB of GON with memory consumption on SMD dataset.}
        \label{fig:hyper}
	\end{minipage}
\end{table}

\textbf{Datasets.} We use two publicly available and 3 self-created datasets in our experiments. We summarize their characteristics in Table~\ref{tab:datasets}.  The \textit{Server Machine Dataset (SMD)} is a five-week long trace of resource utilization of 38 machines from a compute cluster~\cite{omnianomaly}. The \textit{Multi-Source Distributed System (MSDS) Dataset} is a recent high-quality log data composed of distributed traces, application logs, and metrics from a complex distributed system~\cite{nedelkoski2020multi}. We have created another dataset by executing DeFog benchmark applications~\cite{mcchesney2019defog} on a 10-node Raspberry-Pi Cluster. We use an existing fault injection module~\cite{ye2018fault} to create different fault types like CPU overload, RAM contention and DDOS attack using a Poisson process. We truncate the dataset to have only 1, 25 or 55 dimensions and call these FTSAD-1/25/55 datasets (more details in Appendix~\ref{sec:ftsad}).

\textbf{Evaluation Testbed.} We use hyperparameters of the baseline models as presented in their respective papers. For GON, we use batch size $m=64$. We train all models using the PyTorch-1.7.1~\cite{paszke2019pytorch} library. All model training and experiments were performed on a system with configuration: Raspberry-Pi 4B with 8GB RAM and Ubuntu Desktop 20.04 OS.

\begin{figure}[t]
    \centering \setlength{\belowcaptionskip}{-6pt}
    \includegraphics[width=\linewidth]{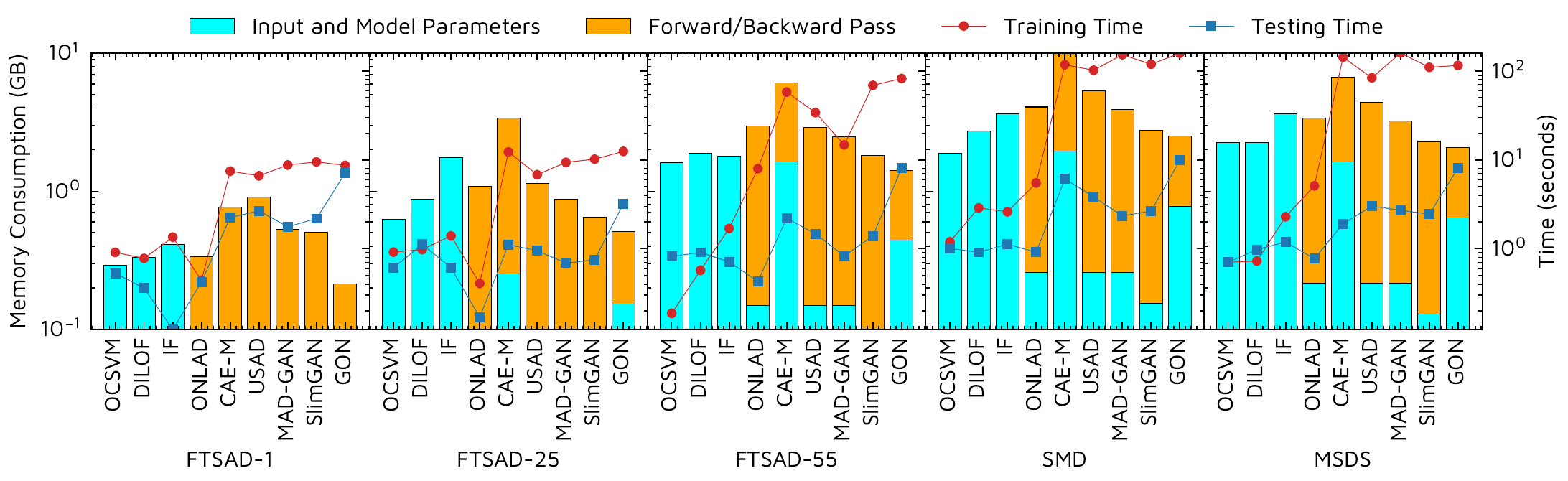}
    \caption{Comparison of the peak memory consumption, training and testing times for all models. The y-axes are in log scale.}
    \label{fig:overhead}
\end{figure}

\textbf{Results.} We compare the F1 score of anomaly prediction as well as the ratio of F1 score with the memory consumption of the model. For each dataset and for each model, we chose the hyperparameters such that F1/GB is maximized. This is because increasing the memory footprint improves anomaly detection accuracy at the cost of the performance of applications running of the edge devices (see Fig.~\ref{fig:hyper} for GON on SMD dataset; more details in Appendix~\ref{sec:hyperparam}). The results are summarized in Table~\ref{tab:results} with additional results in Appendix~\ref{sec:additional} (all values have 95\% confidence bounds of $\leq 0.001$). Among the baselines, the SlimGAN baseline has the highest average F1 score of 0.8227. The GON model gives a higher average F1 score of 0.8784, 6.77\% higher than SlimGAN. When comparing F1/GB, GON improves by 10.41\%-157.09\% with respect to SlimGAN.  Fig.~\ref{fig:overhead} presents the memory consumption, training and testing times for all models. The memory consumption (in GBs) is divided into the size of inputs and network parameters, and the data consumption while executing forward or backward pass. Across all datasets, the memory footprint of GON is the lowest, 9.16\%-57.89\% lower compared to SlimGAN. However, the optimization loop to generate new samples ($z^*$s) has a high overhead, giving 4.98\%-39.67\% higher training times compared to SlimGAN. These results show that having only a single discriminator network allows us to significantly reduce the network's number of parameters and give a performance boost compared to having two such networks.

\begin{table}[t]
\caption{Performance comparison of GON with baseline methods. F1: Macro F1 score, F1/GB: ratio of F1 score to the total memory consumption in GB. The best scores are highlighted in bold.}
    \centering \resizebox{\textwidth}{!}{
    \begin{tabular}{@{}lcccccccccc@{}}
\toprule 
\multirow{2}{*}{Method} & \multicolumn{2}{c}{FTSAD-1} & \multicolumn{2}{c}{FTSAD-25} & \multicolumn{2}{c}{FTSAD-55} & \multicolumn{2}{c}{SMD} & \multicolumn{2}{c}{MSDS}\tabularnewline
\cmidrule{2-11}
 & F1 & F1/GB & F1 & F1/GB & F1 & F1/GB & F1 & F1/GB & F1 & F1/GB\tabularnewline
\midrule 
OCSVM & 0.7047 & 2.4288 & 0.7365 & 1.1806 & 0.7982 & 0.4945 & 0.2938 & 0.1560 & 0.2934 & 0.1301\tabularnewline
DILOF & 0.5989 & 1.7942 & 0.7547 & 0.8614 & 0.8937 & 0.4746 & 0.3134 & 0.1148 & 0.3206 & 0.1420\tabularnewline
IF & 0.6683 & 1.6224 & 0.8107 & 0.4599 & 0.8914 & 0.4982 & 0.2294 & 0.0628 & 0.4326 & 0.1192\tabularnewline
ONLAD & 0.6218 & 1.8430 & \textbf{0.9882} & 0.9114 & 0.9395 & 0.3180 & 0.6227 & 0.1528 & 0.6439 & 0.1899\tabularnewline
CAE-M & 0.7191 & 0.9323 & 0.9786 & 0.2817 & \textbf{0.9495} & 0.1523 & 0.5861 & 0.0572 & 0.6879 & 0.1002\tabularnewline
USAD & 0.7996 & 0.8769 & 0.9724 & 0.8553 & 0.8645 & 0.2991 & 0.7219 & 0.1359 & 0.5312 & 0.1202\tabularnewline
MAD-GAN & 0.7687 & 1.4426 & 0.9663 & 1.1083 & 0.9354 & 0.3793 & 0.6287 & 0.1615 & 0.6100 & 0.1883\tabularnewline
SlimGAN & 0.7521 & 1.4844 & 0.9686 & 1.4926 & 0.9449 & 0.5211 & 0.8214 & 0.2980 & 0.6266 & 0.2731\tabularnewline
GON & \textbf{0.8143} & \textbf{3.8163} & 0.9819 & \textbf{1.9242} & 0.9462 & \textbf{0.6719} & \textbf{0.8238} & \textbf{0.3290} & \textbf{0.8258} & \textbf{0.3963}\tabularnewline
\bottomrule 
\end{tabular}} \vspace{-10pt}
    \label{tab:results}
\end{table}

\section{Conclusions and Future Work}
\label{sec:conclusions}



We present a new framework for memory-efficient generative modeling. Unlike previously proposed GAN models, GON uses a single neural network to both discriminate and generate samples. This comes with the advantage of significant savings in memory footprint, but at the cost of higher training and inference times. Unlike prior work aiming to reduce the parameter size of typical generative models, this ground-up framework is pertinent for memory-constrained systems like edge devices. The training overheads would typically be acceptable in settings with (1) low sample size like few-shot or transfer learning, (2) low data dimension size or (3) time abundant settings like task scheduling or fault-tolerant edge computing where GON might execute in a few seconds, whereas decisions need to be taken every few minutes~\cite{tuli2021cosco}.

The framework allows many extensions. (1) Advances like the conditional generation in GANs. (2) Efficiency improvements in model size (parameters/computational memory footprint) and optimization loop, by devising improved methods for early-stopping in the loop and balancing the trade-off between training performance and convergence speed. (3) Higher-order gradients or gradient approximations for faster convergence. (4) GONs face a similar problem as mode collapse in GANs, when noise samples with the same local optimum generate correlated outputs. Advances in model training and loss formulations could help avoid infrequent optima and increase generation diversity. (5) Using improved starting points by masking the noise space for swift and diverse sample generation.



\section*{Acknowledgements}
Shreshth Tuli is supported by the President's PhD scholarship at Imperial College London. The authors thank Gantavya Bhatt for helpful discussions.

\bibliographystyle{abbrvnat}
{\small
\bibliography{references}}


\appendix

\section{Collecting the FTSAD-1/25/55 Datasets}
\label{sec:ftsad}
The Fog Time Series Anomaly Detection (FTSAD) is a suite of datasets collected on a Raspberry Pi 4B compute cluster. The cluster consists of five 4GB versions and five 8GB versions. We ran the DeFog benchmark applications~\cite{mcchesney2019defog}, specifically the Yolo, PocketSphinx and Aeneas workloads. We use a random scheduler and execute these tasks as Docker containers~\cite{tuli2021cosco}. 
We use an existing fault injection module~\cite{ye2018fault} to create different fault types like CPU overload, RAM contention, Disk attack and DDOS attack. In a CPU overload attack, a simple CPU hogging application is executed that create contention of the compute resources. In RAM contention attack, a program is run that performs continuous memory read/write operations. In disk attack, we run the IOZone benchmark that consumes a large portion of the disk bandwidth. In DDOS attack, we perform several invalid HTTP server connection requests causing network bandwidth contention. We use $Poisson(5)$ distribution with faults being sampled uniformly at random from the attack set. More details are given in~\cite{ye2018fault}.

For the 10-node cluster, we collect traces of resource utilizations including CPU, RAM, disk and network. Each datapoint is collected at the interval of 10 seconds. We truncate the dataset to have only 1, 25 or 55 dimensions and call these FTSAD-1/25/55 datasets. For FTSAD-1, we use the CPU utilization of one of the cluster nodes. For FTSAD-25, we use the CPU and network read bandwidth utilization traces for all 10 nodes and RAM utilization for the five 4GB nodes (as 8GB nodes rarely have memory contentions). For FTSAD-55, we use the CPU utilization, disk read, disk write, network read, network write bandwidths for all nodes and RAM utilization of the five 4GB nodes. The overall traces are collected for 737 minutes, giving 4422 datapoints. For the FTSAD-55 dataset, we split them at the 2158-\textit{th} interval at which we started saving the anomaly labels as well, giving a test set size of 2264 datapoints. For the FTSAD-1 and FTSAD-25, we truncated the datasets to 600 train-5000 test and 574 train-1700 test points respectively. Both datasets were manually curated to have low anomaly frequency in FTSAD-1 and high frequency in FTSAD-25. This allowed us to create a collection of datasets with diverse fault characteristics and dataset sizes.

\section{Hyperparameter Selection Criteria}
\label{sec:hyperparam}

The baseline implementations are directly taken from their public code repositories (after necessary adaptations to work with the Python based environment):
IF\footnote{\url{https://github.com/tharindurb/iNNE}}, DILOF\footnote{\url{http://di.postech.ac.kr/DILOF/}}, CAE-M (with CNN Slimming\footnote{\url{https://github.com/liuzhuang13/slimming}}), USAD\footnote{\url{https://github.com/manigalati/usad}} (with model compression\footnote{\url{https://github.com/antspy/quantized_distillation}}), MAD-GAN\footnote{\url{https://github.com/LiDan456/MAD-GANs}} (with GAN slimming\footnote{\url{https://github.com/VITA-Group/GAN-Slimming}}) and SlimGAN\footnote{\url{https://github.com/houliangict/SlimGAN}}. Other models were developed by us. For training we used learning rate of $10^{-4}$ with weight decay of $10^{-5}$. We used \texttt{Adam} optimizer with \texttt{CosineAnnealing} learning rate decay with warm restarts every 10 epochs. For POT parameters, $\mathrm{coefficient} = 10^{−4}$ for all data sets, low quantile is 0.07 for FTSAD-1/25/55, 0.01 for SMD, and 0.001 for MSDS. These are selected as per~\cite{omnianomaly}. 

\begin{wrapfigure}[]{r}{0.36\textwidth}
    \centering \setlength{\belowcaptionskip}{-9pt}
    \includegraphics[width=\linewidth]{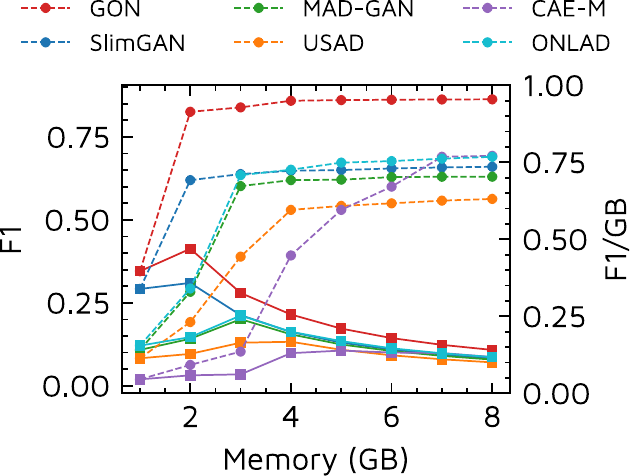}
    \caption{\footnotesize F1 (dotted lines) and F1/GB (solid lines) with memory consumption for all models on the MSDS dataset.}
    \label{fig:hyperall}
\end{wrapfigure}
All baseline methods provide the flexibility of changing the memory footprint of the trained model. In IF we can control the depth and number of trees; in DILOF window size and number of past windows can be controlled; in deep learning models like ONLAD, we can control the neural network size. However, for USAD, CAE-M, MAD-GAN and SlimGAN we can also control the parameter size factor provided by the model compression, CNN/GAN slimming or slimmable layers~\cite{mc, cnn_slimming, gan_slimming, slimgan}. For GON, we only change the number of layers (keeping fixed layer size of 128 nodes). We choose the model memory footprint based on grid search by maximizing F1/GB. The motivation behind this metric is as follows. Increasing memory footprint leads to a monotonic increase in the anomaly detection performance (see Fig.~\ref{fig:hyper}). However, when running online anomaly detection on edge devices, several other applications may be running alongside. Having the maximum possible F1 score would impact negatively the performance of the running applications due to memory contentions. However, having a low memory footprint model but with poor F1 score would be ineffective in detecting anomalies for apt remediation. Thus, a careful balance needs to be maintained to prevent memory contentions and provide high detection. This can be seen by Fig.~\ref{fig:hyper} that shows the F1 and F1/GB scores of GON model on the SMD dataset. For the model with memory consumption of $\sim$2.5GB the F1/GB is maximum and hence is used in our experiments. Similarly, we determine the compression and slimming factors in ONLAD, USAD, CAE-M, MAD-GAN and SlimGAN (see Fig.~\ref{fig:hyperall} for the MSDS dataset).

\section{Additional Results}
\label{sec:additional}

\begin{table}[]
    \centering \renewcommand{\arraystretch}{0.95}
    \caption{Performance comparison of GON with baseline methods. P: Precision, R: Recall, F1: Macro F1 score, ROC/AUC: Area under the ROC curve, Memory: RAM consumption in GB, F1/GB: ratio of F1 score to the total memory consumption in GB, Train/Test time: time taken in seconds to run training or prediction on the test set. The best scores are highlighted in bold.}
    \resizebox{\textwidth}{!}{
\begin{tabular}{@{}lcccccccc@{}}
\toprule
\multirow{2}{*}{Method} & \multicolumn{8}{c}{FTSAD-1}\tabularnewline
\cmidrule{2-9}
 & P & R & F1 & ROC/AUC & Memory & F1/GB & Train Time & Test Time\tabularnewline
\midrule 
OCSVM & 0.8292 & 0.6128 & 0.7047 & 0.7361 & 0.2902 & 2.4288 & 0.9128 & 0.5318\tabularnewline
DILOF & 0.6848 & 0.5321 & 0.5989 & 0.7214 & 0.3338 & 1.7942 & 0.7812 & 0.3617\tabularnewline
IF & 0.7655 & 0.5929 & 0.6683 & 0.7383 & 0.4119 & 1.6224 & 1.3531 & \textbf{0.1239}\tabularnewline
ONLAD & 0.5672 & 0.6879 & 0.6218 & 0.8051 & 0.3374 & 1.8430 & \textbf{0.4428} & 0.4233\tabularnewline
CAE-M & 0.7442 & 0.6957 & 0.7191 & 0.8301 & 0.7713 & 0.9323 & 7.4859 & 2.2560\tabularnewline
USAD & 0.9264 & 0.7034 & 0.7996 & 0.8476 & 0.9120 & 0.8769 & 6.6316 & 2.6814\tabularnewline
MAD-GAN & 0.8385 & \textbf{0.7096} & 0.7687 & 0.8447 & 0.5329 & 1.4426 & 8.7763 & 1.7707\tabularnewline
SlimGAN & 0.8039 & 0.7065 & 0.7521 & 0.8405 & 0.5066 & 1.4844 & 9.5259 & 2.1909\tabularnewline
GON & \textbf{0.9977} & 0.6879 & \textbf{0.8143} & \textbf{0.8438} & \textbf{0.2134} & \textbf{3.8163} & 8.6917 & 7.1232\tabularnewline
\midrule 
\multirow{2}{*}{Method} & \multicolumn{8}{c}{FTSAD-25}\tabularnewline
\cmidrule{2-9}
 & P & R & F1 & ROC/AUC & Memory & F1/GB & Train Time & Test Time\tabularnewline
\midrule 
OCSVM & 0.8713 & 0.6378 & 0.7365 & 0.5632 & 0.6238 & 1.1806 & 0.9189 & 0.6123\tabularnewline
DILOF & 0.8213 & 0.6981 & 0.7547 & 0.5314 & 0.8761 & 0.8614 & 0.9812 & 1.1230\tabularnewline
IF & 0.9173 & 0.7264 & 0.8107 & 0.5324 & 1.7630 & 0.4599 & 1.3976 & 0.6125\tabularnewline
ONLAD & \textbf{0.9803} & 0.9962 & \textbf{0.9882} & 0.9852 & 1.0842 & 0.9114 & \textbf{0.4081} & \textbf{0.1698}\tabularnewline
CAE-M & 0.9580 & 1.0000 & 0.9786 & 0.9896 & 3.3920 & 0.2817 & 12.3181 & 1.1123\tabularnewline
USAD & 0.9580 & 0.9871 & 0.9724 & 0.9896 & 1.1369 & 0.8553 & 6.8215 & 0.9566\tabularnewline
MAD-GAN & 0.9547 & 0.9781 & 0.9663 & 0.9883 & 0.8718 & 1.1083 & 9.4016 & 0.6903\tabularnewline
SlimGAN & 0.9497 & 0.9882 & 0.9686 & 0.9883 & 0.6489 & 1.4926 & 10.2125 & 0.7531\tabularnewline
GON & 0.9665 & \textbf{0.9978} & 0.9819 & \textbf{0.9918} & \textbf{0.5103} & \textbf{1.9242} & 12.5008 & 3.2312\tabularnewline
\midrule 
\multirow{2}{*}{Method} & \multicolumn{8}{c}{FTSAD-55}\tabularnewline
\cmidrule{2-9} 
 & P & R & F1 & ROC/AUC & Memory & F1/GB & Train Time & Test Time\tabularnewline
\midrule 
OCSVM & 0.8548 & 0.7486 & 0.7982 & 0.5623 & 1.6141 & 0.4945 & \textbf{0.1879} & 0.8192\tabularnewline
DILOF & 0.8721 & 0.9163 & 0.8937 & 0.5887 & 1.8829 & 0.4746 & 0.5730 & 0.9128\tabularnewline
IF & 0.8913 & 0.8916 & 0.8914 & 0.5446 & 1.7892 & 0.4982 & 1.6958 & 0.7124\tabularnewline
ONLAD & 0.9038 & 0.9782 & 0.9395 & 0.9252 & 2.9542 & 0.3180 & 7.9839 & \textbf{0.4304}\tabularnewline
CAE-M & 0.9038 & 1.0000 & 0.9495 & \textbf{0.9916} & 6.0870 & 0.1523 & 58.1240 & 2.2133\tabularnewline
USAD & 0.7673 & 0.9898 & 0.8645 & 0.9928 & 2.8905 & 0.2991 & 34.1498 & 1.4748\tabularnewline
MAD-GAN & \textbf{0.9530} & 0.9183 & 0.9354 & 0.9252 & 2.4663 & 0.3793 & 14.7811 & 0.8315\tabularnewline
SlimGAN & 0.9038 & 0.9899 & 0.9449 & \textbf{0.9916} & 1.8131 & 0.5211 & 69.1473 & 1.3958\tabularnewline
GON & 0.9038 & \textbf{0.9928} & \textbf{0.9462} & \textbf{0.9916} & \textbf{1.4083} & \textbf{0.6719} & 82.5418 & 8.1239\tabularnewline
\midrule 
\multirow{2}{*}{Method} & \multicolumn{8}{c}{SMD}\tabularnewline
\cmidrule{2-9} 
 & P & R & F1 & ROC/AUC & Memory & F1/GB & Train Time & Test Time\tabularnewline
\midrule 
OCSVM & \textbf{0.9812} & 0.1728 & 0.2938 & 0.5623 & 1.8831 & 0.1560 & \textbf{1.1984} & 1.0128\tabularnewline
DILOF & 0.9643 & 0.1871 & 0.3134 & 0.5887 & 2.7307 & 0.1148 & 2.8919 & \textbf{0.9123}\tabularnewline
IF & 0.9271 & 0.1309 & 0.2294 & 0.5446 & 3.6536 & 0.0628 & 2.6141 & 1.1294\tabularnewline
ONLAD & 0.4604 & 0.9617 & 0.6227 & 0.9018 & 4.0741 & 0.1528 & 5.5175 & 0.9141\tabularnewline
CAE-M & 0.4145 & \textbf{1.0000} & 0.5861 & \textbf{0.9098} & 10.0108 & 0.0572 & 118.1238 & 6.1291\tabularnewline
USAD & 0.5774 & 0.9627 & 0.7219 & 0.9028 & 5.3099 & 0.1359 & 102.1547 & 3.8555\tabularnewline
MAD-GAN & 0.4604 & 0.9912 & 0.6287 & 0.8874 & 3.8921 & 0.1615 & 153.1956 & 2.3478\tabularnewline
SlimGAN & 0.7928 & 0.8522 & 0.8214 & 0.9001 & 2.7568 & 0.2980 & 119.4933 & 2.6660\tabularnewline
GON & 0.8439 & 0.8047 & \textbf{0.8238} & 0.8928 & \textbf{2.5042} & \textbf{0.3290} & 158.9011 & 9.9102\tabularnewline
\midrule 
\multirow{2}{*}{Method} & \multicolumn{8}{c}{MSDS}\tabularnewline
\cmidrule{2-9} 
 & P & R & F1 & ROC/AUC & Memory & F1/GB & Train Time & Test Time\tabularnewline
\midrule 
OCSVM & 0.9722 & 0.1728 & 0.2934 & 0.5445 & 2.2560 & 0.1301 & \textbf{0.7065} & \textbf{0.7065}\tabularnewline
DILOF & 0.9891 & 0.1913 & 0.3206 & 0.6445 & 2.2581 & 0.1420 & 0.7277 & 0.9820\tabularnewline
IF & 0.9359 & 0.2813 & 0.4326 & 0.5619 & 3.6300 & 0.1192 & 2.2997 & 1.1928\tabularnewline
ONLAD & 0.5873 & 0.7126 & 0.6439 & 0.8319 & 3.3903 & 0.1899 & 5.1128 & 0.7747\tabularnewline
CAE-M & 0.5243 & \textbf{1.0000} & 0.6879 & \textbf{0.9214} & 6.7052 & 0.1002 & 143.1294 & 1.9128\tabularnewline
USAD & 0.3616 & \textbf{1.0000} & 0.5312 & 0.7913 & 4.4187 & 0.1202 & 84.0544 & 3.0318\tabularnewline
MAD-GAN & 0.4778 & 0.8434 & 0.6100 & 0.8927 & 3.2389 & 0.1883 & 159.7400 & 2.7163\tabularnewline
SlimGAN & \textbf{0.9982} & 0.4566 & 0.6266 & 0.5783 & 2.2941 & 0.2731 & 110.3777 & 2.4567\tabularnewline
GON & 0.8089 & 0.8434 & \textbf{0.8258} & 0.9154 & \textbf{2.0839} & \textbf{0.3963} & 115.8802 & 8.1239\tabularnewline
\bottomrule 
\end{tabular}}
    \label{tab:additionalresults}
\end{table}

Table~\ref{tab:additionalresults} presents additional results including precision, recall, F1 score, memory consumption, area under the receiver operating characteristic (ROC) curve, F1/GB, training and testing times. In summary, the results show that the detection accuracy of GON is higher than the baseline methods at the cost of higher training and testing times. Compared to the most accurate baseline, SlimGAN, GON has up to 24.11\%, 84.71\% and 58.29\% higher precision, recall and ROC/AUC. For even tighter memory constraints (< 8GB RAM), the performance improvement with GON is higher, but we present results for 8GB limit as this is minimum memory size in commonplace edge devices. For distributed edge computing environments, where tasks like anomaly detection need to run online in memory constrained devices, GON is the \textit{best} method in terms of the memory footprint. The memory consumption of GON is up to 57.88\% lower than SlimGAN, preventing the execution of the GON model itself from causing memory contentions for other computational tasks. For instance, if an edge cluster is running with 10 nodes (like the FTSAD-55 dataset assumption), the CAE-M model is bound to consume $\sim$6GB of memory, leaving up to 2GB RAM for other computational and background tasks running in edge devices. This would have a detrimental effect of the Quality of Service (QoS) of the running applications and more exhaustive experimentation is left as part of future work.

\section{Broader Impact}
\label{sec:broader_impact}

\begin{figure}
    \centering
    \includegraphics[width=\linewidth]{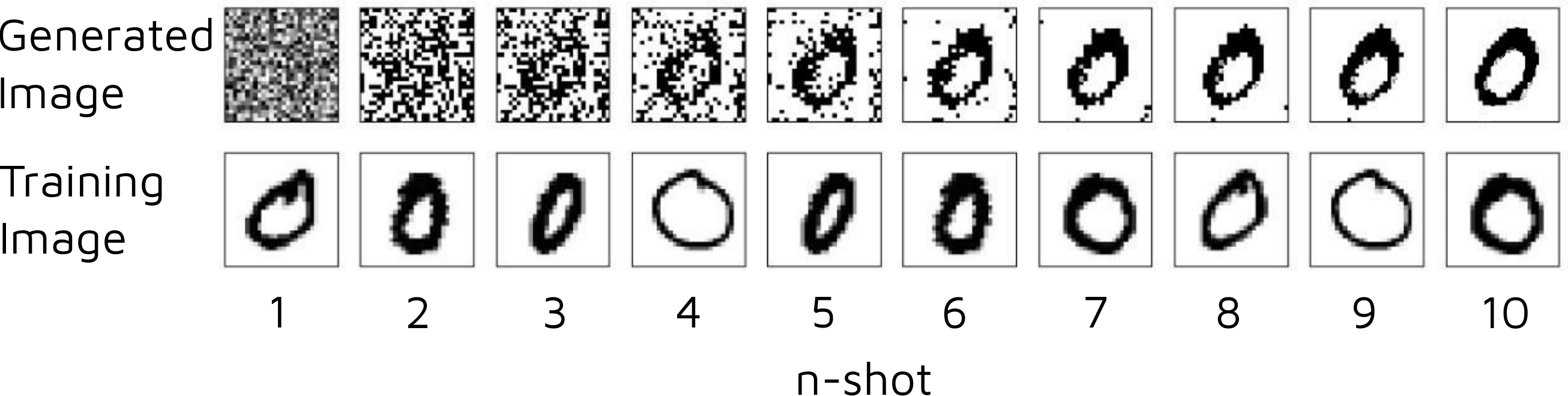}
    \caption{Few-shot handwritten digit generation using GON.}
    \label{fig:mnist}
\end{figure}

\begin{figure}
    \centering
    \includegraphics[width=\linewidth]{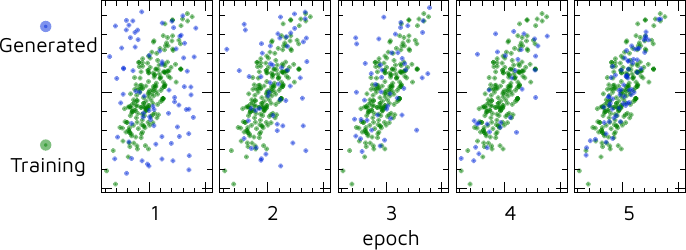}
    \caption{Generating a two-dimensional Gaussian distribution using GON.}
    \label{fig:gauss}
\end{figure}

As described in Section~\ref{sec:intro}, the GON framework is agnostic to the specific problem of anomaly detection and can be used in general for any generation task providing significant savings in memory footprint. As pointed out in Section~\ref{sec:conclusions}, due to the relatively high training times of the framework, it is particularly suitable for generative tasks with low dataset sizes (like few-shot learning) or low dimensional data. To this end, we provide anecdotal evidence of the applicability of GON in two such scenarios.\footnote{Code available at \url{https://github.com/imperial-qore/GON_MNIST}.}

\textbf{Few-shot handwritten digit generation.} Fig.~\ref{fig:mnist} shows how a GON model learns to generate handwritten digits using only 10 samples from the dataset. The generated image after each new sample from the dataset (with replacement) is shown left to right in Fig.~\ref{fig:mnist}. We show with only a single digit (zero in this case) for simplicity. We use the MNIST dataset to train this model~\cite{deng2012mnist}. The generated images start from the same noise samples to highlight the progressive learning. While running optimization in the input space, after each update step, we clip $z$ in the range $[0, 1]$ to keep within the gray-scale pixel range. The complete training procedure took $\sim$10 seconds on the same setup as explained in Section~\ref{sec:experiments}, showing how the GON framework can be easily used in settings with limited training data. GON gives nearly 60\% memory savings with $\sim$27\% higher training time compared to a similar performance GAN model.

\textbf{Generating 2D Gaussian distribution.} Fig.~\ref{fig:gauss} shows how a GON model learns to generate samples from a two-dimensional Gaussian distribution, with complete training (5 epochs) taking $\sim$3 seconds. The green points show the training data and the blue points show the samples generated by the GON model.  Within five epochs, the model is able to effectively replicate the underlying distribution of the data. The figure illustrates that the GON model generates diverse samples. Moreover, it shows how GON can be reasonably trained for density estimation of low dimensional datasets. Again, GON gives 82\% memory savings with only $\sim$7\% higher training times.

\end{document}